\documentclass[10pt,twocolumn,letterpaper]{article}

\usepackage[pagenumbers]{cvpr} 

\usepackage{graphicx}
\usepackage{amsmath}
\usepackage{amssymb}
\usepackage{booktabs}
\usepackage{setspace}

\usepackage[pagebackref,breaklinks,colorlinks]{hyperref}

\usepackage[capitalize]{cleveref}
\crefname{section}{Sec.}{Secs.}
\Crefname{section}{Section}{Sections}
\Crefname{table}{Table}{Tables}
\crefname{table}{Tab.}{Tabs.}

\def\cvprPaperID{*****} 
\def\confName{CVPR}
\def\confYear{2022}

\newcommand\blfootnote[1]{%
    \begingroup
    \renewcommand\thefootnote{}\footnote{#1}%
    \addtocounter{footnote}{-1}%
    \endgroup
}

\begin{document}

\title{DETR++: Taming Your Multi-Scale Detection Transformer}

\author{
    Chi Zhang$^{1,2,^{\star}}$, Lijuan Liu$^2$, Xiaoxue Zang$^{2,\star}$, \\ Frederick Liu$^2$, Hao Zhang$^2$, Xinying Song$^2$, Jindong Chen$^2$\\
    $^1$ Department of Computer Science, University of California, Los Angeles\\
    $^2$ Google Research\\
    {\tt\small chi.zhang@ucla.edu, lijuanliu@google.com, zxx1204007@gmail.com}\\ 
    {\tt\small \{frederickliu,haozhangthu,xysong,jdchen\}@google.com}
}
\maketitle

\begin{abstract}
Convolutional Neural Networks (CNN) have dominated the field of object detection ever since the success of AlexNet in ImageNet classification~\cite{krizhevsky2012imagenet}. With the sweeping reform of Transformers~\cite{vaswani2017attention} in natural language processing, Carion \etal~\cite{carion2020end} introduce the Transformer-based detection method, \ie, DETR. However, due to the quadratic complexity in the self-attention mechanism in the Transformer, DETR is never able to incorporate multi-scale features as performed in existing CNN-based detectors, leading to inferior results in small object detection. To mitigate this issue and further improve performance of DETR, in this work, we investigate different methods to incorporate multi-scale features and find that a Bi-directional Feature Pyramid (BiFPN) works best with DETR in further raising the detection precision. With this discovery, we propose \emph{DETR++}, a new architecture that improves detection results by $\mathbf{1.9\%}$ AP on MS COCO 2017, $\mathbf{11.5\%}$ AP on RICO icon detection, and $\mathbf{9.1\%}$ AP on RICO layout extraction over existing baselines.
\end{abstract}

\setstretch{0.99}
\section{Introduction}
\label{sec:intro}

\blfootnote{$^\star$ work done while at Google Research.}

All well-performing existing object detection systems leverage the Convolutional Neural Networks (CNN) and address the set prediction problem in matching box proposals and ground truths using predefined heuristics: typical examples include anchors~\cite{girshick2014rich,girshick2015fast,ren2015faster,lin2017focal,liu2016ssd}, grid~\cite{redmon2016you,redmon2017yolo9000,redmon2018yolov3}, point centers~\cite{zhou2019objects,tian2019fcos}. However, while good practices have been identified in widely used datasets, like MS COCO~\cite{lin2014microsoft} and PASCAL~\cite{everingham2015pascal}, the post-processing steps are critical performance factors for relatively under-explored areas, like icon detection and layout extraction in on-device screen understanding tasks~\cite{deka2017rico}.

A new Transformer-based~\cite{vaswani2017attention} detection method, \ie, DETR~\cite{carion2020end}, has been recently proposed to mitigate these issues. Specifically, the DETR model uses a ResNet~\cite{he2016deep} backbone to extract higher-level visual features. A Transformer encoder further aggregates global features in each token from the flattened visual features. Finally, a Transformer decoder decodes the encoded features into box proposals. While the backbone and the encoder are relatively standard, the design in the decoder does make a difference. Specifically, the decoder module takes as input a zero matrix as the sentence embedding and learnable position embedding referred to as \emph{object queries}. Position embedding for the encoder and the object queries are added to each attention layer's key and query respectively and decoded parallelly. A final object classifier and a box regressor are attached to the output features of the decoder and generate object proposals represented as $(c, cy, cx, h, w)$, denoting the class $c$, box center $(cy, cx)$, and box height and width $(h, w)$. The box proposals and the ground truth boxes are matched via the Hungarian algorithm, relieving the detector designer from crafting the matching heuristics, and the classification and box regression losses are jointly minimized.

Despite the simpler design in the DETR architecture, earlier experimental results show that the DETR model is inferior to existing convolutional models and also slower in training. There are two sources contributing to complexity in the model: (1) the self-attention mechanism in the encoder is resource-hungry, especially for visual features that could span over thousands of tokens and (2) the Hungarian matcher is cubic in time. These slow operations make the common strategy of adding multi-scale features in a detector to improve performance a non-trivial work: running the detector head on multi-scale features, \eg, $160\times160$ in the visual feature hierarchy, or simply increasing the number of proposals is extremely memory- and time-consuming.

Therefore, in this work, we set out to study what is the best way to incorporate multi-level features into the DETR architecture to improve performance, while not incurring the quadratic complexity and the cubic complexity in the self-attention and the matcher. Specifically, we note that: (1) running the head on multi-level features is nearly impossible given the resource and time constraints, (2) the Transformer encoder plays a crucial role in the detector and cannot be removed, (3) the shifted window idea~\cite{liu2021swin} does not work well on multi-level features, (4) specialized DETR heads for different object scales linearly increase complexity but do not fare better than the baseline, (5) a Bi-directional Feature Pyramid architecture improves the performance and only marginally adds to the complexity.

In the remainder of this paper, we review existing works on object detection and multi-level features in \cref{sec:related}, discuss the multi-scale designs we explored and introduce DETR++ in \cref{sec:model}, detail the experimental setup in \cref{sec:exp}, and finally conclude the paper in \cref{sec:conclude}.

\section{Related Work}
\label{sec:related}

\subsection{Object Detectors}

Common object detection systems can be categorized into two routes: two-stage object detection and one-stage object detection. In the two-stage detectors, a region proposal network will first propose potential bounding boxes that may contain objects and a final classifier decides what kind of an object the bounding box contains. The most typical two-stage object detector is the Faster R-CNN~\cite{ren2015faster}, whose Region Proposal Network and the classifier head are responsible for the two sub-tasks respectively. The one-stage object detectors merge the two sub-routines into one by either predefining the proposal grids~\cite{redmon2016you}, the object anchors~\cite{liu2016ssd,lin2017focal,tian2019fcos}, or the dense object center points~\cite{zhou2019objects}. These detection systems predict corresponding bounding boxes with respect to the predefined constructs from the same features used for the object class prediction. However, these one-stage systems fare slightly worse than the two-stage systems potentially due to the region mismatch that is better handled by a region proposal stream. However, neither the two-stage detectors nor the one-stage detectors are relieved from the manual post-processing step of Non-Maximum Suppression (NMS).

A recently proposed Transformer-based detector, \ie, DETR, successfully removes the post-processing steps using set-based matching and prediction. In particular, this one-stage detector predicts bounding boxes and object classes parallelly from an encoder-decoder architecture and matches the proposals with ground truths via Hungarian matching~\cite{jonker1987shortest}. The matched boxes incur the classification loss and box regression loss, whose gradients are backpropagated through the entire model.

\subsection{Multi-Scale Detection}

One critical component of existing object detection systems is multi-level feature aggregation. Multi-level feature aggregation allows the model to leverage features of different granularity and would effectively improve the model on small object detection. The most popular multi-scale strategy is using Feature Pyramid Network (FPN)~\cite{lin2017feature}. Specifically, multi-level features from the detection backbone are processed through FPN: the resolution of each level remains the same and the detection head is attached onto feature maps of all resolutions. The detection results are then NMS-filtered to produce the final prediction. Following the idea, the PANet~\cite{liu2018path} method supplements the top-down direction of FPN with a bottom-up path. Ghiasi \etal~\cite{ghiasi2019fpn} even consider using Neural Architecture Search to automatically find a better FPN structure. Recently, EfficientDet~\cite{tan2020efficientdet} proposes a simpler, scalable, and effective FPN module that employs the bi-directional idea, residual connection and weighted averaging.

However, due to the computational complexity in DETR, the original model cannot easily incorporate a multi-scale feature aggregation module. The lack of multi-level feature representation negatively impacts small objects' detection precision in the original DETR model.

\section{Multi-Scale Designs}
\label{sec:model}


In this section, we discuss a few potential multi-scale strategies before finally presenting how we incorporate an effective design into the original model to make DETR++.

With the time complexity of the self-attention mechanism in the Transformer encoder being $O(N^2)$ and that of the Hungarian matching being $O(N^3)$, simply using large-resolution features for detection and (or) increasing the number of box predictions are not computationally feasible. Therefore, we investigate the following strategies to incorporate multi-scale features.

\begin{figure}[t!]
    \centering
    \includegraphics[width=\linewidth]{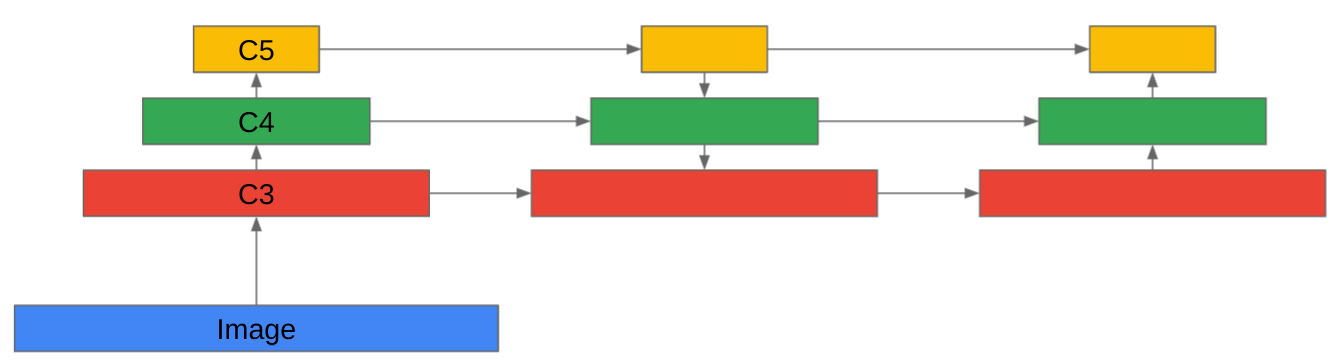}
    \caption{The idea of Bi-direction Feature Pyramid, where, apart from the top-down and lateral pathway in the traditional Feature Pyramid, an additional bottom-up pathway is added.}
    \label{fig:bifpn}
\end{figure}

\subsection{Removing the Encoder}

Now that one of the computational burdens is from the self-attention mechanism, one intuitive idea is to remove the encoder and only use the decoder as the decoder's complexity is only of $O(MN)$, where $M$ denotes the number of object queries ($100$ in DETR) and $N$ denotes the size of the encoder output.

We consider two potential aggregation methods using the decoder only.

\paragraph{Stack} In this stacking strategy, we consecutively apply three decoders on the image features from $C_3$, $C_4$, and $C_5$. The decoded output from $C_3$ is further processed by the decoder for $C_4$, followed by the $C_5$ decoder. We use six-layer transformer decoders for the three scales of input and compute the auxiliary loss on every layer's output.

\paragraph{Multi-Head} We also consider the multi-head method, where we use three six-layer decoders for each resolution similarly to the stacking method. However, unlike the stacking method, each decoder independently produces the box proposals from a single scale. For loss computation, we concatenate all box proposals from all the three decoders.

\subsection{Shifted Windows}

Inspired by the Swin Transformer~\cite{liu2021swin}, we consider applying the Transformer detection head on each shifted window. This will only linearly increase computation rather than quadratically. 

In particular, we set the window size to be the spatial size of the $C_5$ features and correspondingly crop non-overlapping patches out of $C_4$ and (or) $C_3$ features. The patches are sent to the Transformer-based detection head and the proposal boxes are concatenated. In practice, we only use $C_4$ and $C_5$ since adding two many patches will correspondingly increase the number of boxes, which slows down training due to the cubic complexity in Hungarian matching. 

\subsection{Specialized Heads}

Another way to better leverage multi-scale information, however, is to disentangle detection for objects of different scales, such that features from larger-scale objects will not negatively impact small-scale objects and vice verse. To implement the idea, we propose to use specialized heads for object detection of different scales.

Specifically, we use three detection heads for small objects, medium objects, and large objects, respectively. All the detection heads operate on the $C_5$ features and produce box proposals to be concatenated for loss computation. 

In practice, we split the object boxes in data processing into the three scales, and during training, we use separate Transformer detection heads for each scale.

\subsection{Bi-directional Feature Pyramid}

Conventionally, the Feature Pyramid Network~\cite{lin2017feature} is used to produce multiple levels of features, each of which is attached to the detection head for object detection. However, this method simply cannot be used due to the quadratic complexity in the Transformer encoder. Therefore, we propose to aggregate the multiple features using a Bi-directional Feature Pyramid Network (BiFPN)~\cite{tan2020efficientdet}. BiFPN works by supplying the top-down direction of the traditional FPN with an additional bottom-up pathway. See \cref{fig:bifpn} for a graphical illustration of the idea. We also consider stacking BiFPN layers to produce better representation. Finally, we pick one scale of the final BiFPN layer's features and attach it to the Transformer detection head for object detection.

\begin{table*}[th!]
    \begin{subtable}[th!]{0.58\textwidth}
        \centering
        \resizebox{\textwidth}{!}{
        \begin{tabular}{l c c c c c c}
            \toprule
            Method               & AP   & AP@0.5 & AP@0.75 & AP$^L$ & AP$^M$ & AP$^S$ \\
            \midrule
            DETR-NoEnc-Stack     & 37.3 & 56.8   & 39.7    & 54.4   & 40.6   & 16.9   \\
            DETR-NoEnc-MHead     & 35.0 & 54.9   & 36.3    & 52.0   & 37.5   & 14.6   \\
            DETR-Swin            & 39.9 & 59.8   & 42.2    & 57.9   & 43.6   & 18.4   \\
            DETR-SHead           & 36.4 & 54.0   & 39.2    & 54.7   & 39.5   & 15.1   \\
            DETR++               & \textbf{41.8} & \textbf{60.1} & \textbf{44.6} & \textbf{58.6} & \textbf{45.0} & 22.1 \\
            \midrule
            DETR                 & 39.9 & 59.8   & 42.4    & 57.2   & 43.3   & 18.8   \\
            CenterNet            & 41.6 & 59.4   & 44.2    & 54.1   & 43.1   & \textbf{22.5} \\
            \bottomrule
        \end{tabular}
        }
        \caption{Model performance on MS COCO 2017, where NoEnc denotes the design without the encoder, MHead is short for multi-head, Swin for shifted windows, and SHead for specialized heads.}
        \label{tbl:coco}
    \end{subtable}
    \hfill
    \begin{subtable}[th!]{0.42\textwidth}
        \centering
        \resizebox{\textwidth}{!}{
        \begin{tabular}{l c c c c c c}
            \toprule
            Method               & AP   & AP@0.5 & AP@0.75 & AP$^L$ & AP$^M$ & AP$^S$ \\
            \midrule
            DETR++               & \textbf{48.1} & \textbf{89.8} & \textbf{45.3} & \textbf{52.9} & \textbf{49.6} & \textbf{43.6} \\
            \midrule
            DETR                 & 47.4 & 89.4   & 44.3    & 52.0   & 48.8   & 43.1   \\
            IconNet              & 36.6 & 79.3   & 26.8    & 15.1   & 35.6   & 36.8   \\
            \bottomrule
        \end{tabular}
        }
        \\
        \resizebox{\textwidth}{!}{
        \begin{tabular}{l c c c c c c}
            \toprule
            Method               & AP   & AP@0.5 & AP@0.75 & AP$^L$ & AP$^M$ & AP$^S$ \\
            \midrule
            DETR++               & \textbf{25.3} & \textbf{43.6} & \textbf{24.2} & \textbf{28.6} & \textbf{9.7} & 1.4 \\
            \midrule
            DETR                 & 24.7 & 42.5   & 23.4    & 28.0   & 8.1    & 1.1    \\
            IconNet              & 16.2 & 30.5   & 15.9    & 19.8   & 8.6    & \textbf{5.6}    \\
            \bottomrule
        \end{tabular}
        }
        \caption{Model performance on RICO icon detection (top) and RICO layout detection (bottom).}
        \label{tbl:RICO}
    \end{subtable}
    \caption{Experimental evaluation.}
\end{table*}

\subsection{DETR++}

\begin{figure}[t!]
    \centering
    \includegraphics[width=\linewidth]{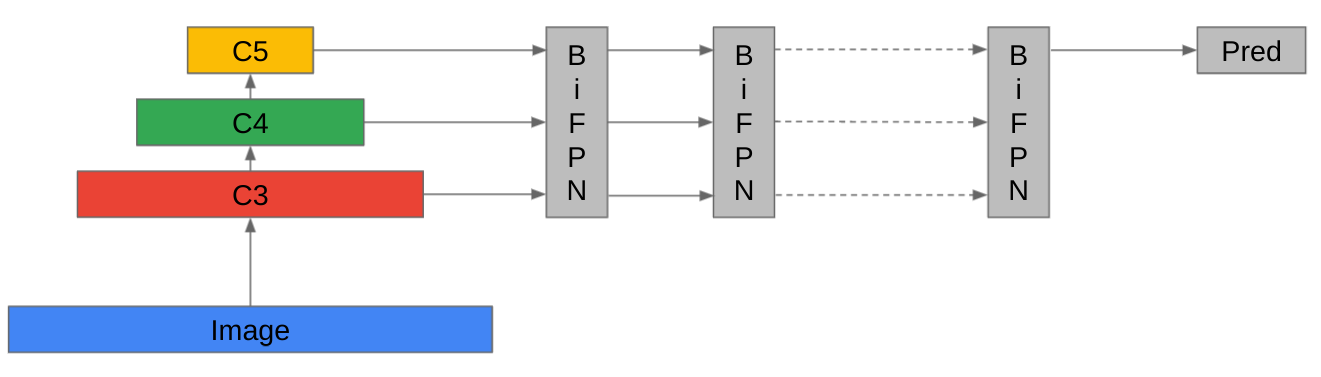}
    \caption{DETR++ model, where multi-scale features are fed into repeated BiFPN layers and only one scale of image features is send to the encoder and decoder (not shown) for detection.}
    \label{fig:detr++}
\end{figure}

After extensive experimentation, we propose the DETR++ architecture that adds the Bi-directional Feature Pyramid into the original DETR model. Specifically, we connect the BiFPN module to the feature output of $C_3$, $C_4$, and $C_5$ from the ResNet backbone, and stack it 8 times, before finally feeding the multi-scale feature-aggregated $C_5$ output to the Transformer architecture. \cref{fig:detr++} shows the final architecture of our model.

\section{Experiments}
\label{sec:exp}

We show DETR++ achieves significant performance improvement over baselines in object detection for natural images and on-device screen understanding tasks like icon detection and layout extraction. 

\paragraph{Training Details} We train DETR++ with the AdamW~\cite{loshchilov2017decoupled} optimizer. The initial learning rate and the weight decay are both set to $1\times10^{-4}$. The learning rate is scheduled to decrease 10 times after 200k steps and the training process completes after 500k steps. We use the ImageNet-pretrained ResNet-50~\cite{he2016deep} as our backbone. Of note, the backbone has the same learning rate with the rest of the model. The 8-layer BiFPN module is attached to the last three intermediate features and we use six layer of blocks in both the Transformer encoder and the decoder. Both the Transformer encoder and the decoder have a dropout~\cite{srivastava2014dropout} rate of 0.1 and apply post normalization.

We use data augmentation in preprocessing. In particular, we use random horizontal flip and random crop during training such that the short size of the image is at least 480 and at most 800 and the long size of the image does not exceed 1280. The image is then padded to be of shape 1280$\times$1280.

We follow the original prediction setup in DETR~\cite{carion2020end}, \ie, we overwrite predictions of background with the most-likely non-background objects and their corresponding confidence.


Original hyperparameter settings have been inherited. However, we note that the number of boxes and the background class weight are two important hyperparameters to tune for different datasets. One rule of thumb we note is to set the background class weight to be approximately
\begin{equation}
    \frac{\text{Ground truth boxes}}{\text{Box proposals} \times \text{Images} - \text{Ground truth boxes}}.
\end{equation}

\paragraph{Datasets} In the experiments, we validate the performance of our model on three datasets, the MS COCO 2017 dataset~\cite{lin2014microsoft} as a standard benchmark for object detection, and RICO icon detection dataset and RICO layout extraction dataset~\cite{deka2017rico} for smart device screen understanding. The COCO dataset contains natural images with annotated common objects; the RICO icon dataset consists of Android device snapshots with annotated icons; the RICO layout dataset provides bounding boxes and class labels for functional areas in Android device screen snapshots.

\paragraph{Performance on MS COCO}

\cref{tbl:coco} shows the performance of various multi-scale schemes on the MS COCO 2017 dataset. As indicated in the table, the idea to remove the encoder will negatively impact the model performance: AP values significantly drop in both the stack design and the multi-head design. The shifted window idea achieves similar performance with the original DETR model with slightly inferior small object detection precision. The specialized head design neither improves over the baseline DETR model but only linearly increases the computation. Compared to others, our DETR++ model with the BiFPN module significantly improves over the baseline DETR model and is even better than the CenterNet model of similar compute~\cite{duan2019centernet} except in small object detection. 

\paragraph{Performance on Icon}

Given the results in MS COCO detection, we evaluate the best mult-scale strategy of BiFPN on the RICO icon detection dataset~\cite{deka2017rico}. \cref{tbl:RICO} shows the results of the experiments. Notably, the DETR++ model improves the IconNet model~\cite{he2020actionbert,bai2021uibert} by a large margin and also fares better than the DETR baseline. DETR++ is better than DETR in every aspect, and achieves even better results than IconNet on small and medium object detection.

\paragraph{Performance on Layout}

The same set of models is run on the RICO layout extraction dataset~\cite{deka2017rico} as well. Consistent with earlier results, the DETR++ model improves the DETR model on every precision metrics. However, in the layout extraction task, the small object detection is more serious than earlier datasets and the DETR++ model becomes inferior to the IconNet in AP$^S$.

\section{Conclusion and Future Work}
\label{sec:conclude}

In this work, we investigate the multi-scale strategies in scaling the DETR model and propose the DETR++ model. Specifically, the DETR++ model incorporates a Bi-direction Feature Pyramid Network to aggregate multi-level image features to improve small object detection in the original DETR model. In experiments, we note that DETR++ improves detection results by 1.9\% AP on MS COCO 2017, 11.5\% AP on icon detection, and 9.1\% AP on layout extraction over existing baselines.




Despite success in using BiFPN for multi-level feature aggregation, there is still room for improvement for DETR++ in small object detection. Specifically, the DETR++ model is still not as performant as CenterNet and IconNet in small object detection in specific datasets. These gaps further motivate us in pursuing this direction to make Transformer-based detectors a first-class citizen.

Convergence speed of the DETR++ model is also slower than existing baselines. This slow-down significantly impacts model iteration. While we have noticed recent attempts to improve convergence speed, few of them will bring the model to the optimal point from the plain method. 

Finally, we see a bigger picture of the Transformer-based detector in multi-modal reasoning~\cite{zhang2019raven,zhang2019learning,zhang2020machine,zhu2020dark,zhang2021abstract,zhang2021acre,zhang2021learning}. With the effective large language learning models, we could combine language features and the Transformer detector to enable open-domain object detection. The direction could potentially change the landscape of detection: the detector is no longer fixed on a domain but could quickly use language features to determine where an object is and what that object is.
\setstretch{1.0}
{\small
\bibliographystyle{ieee_fullname}
\bibliography{egbib}
}

\end{document}